\documentclass[runningheads]{llncs}
\usepackage[T1]{fontenc}
\usepackage{graphicx}
\usepackage{booktabs}
\usepackage[misc]{ifsym}

\usepackage{mwe}
\usepackage{url}

\begin{document}

\title{How well does Classification Accuracy capture Concept Drift Detection Quality? An overview of~Concept Drift Detection evaluation}

\titlerunning{An overview of~Concept Drift Detection evaluation}

\author{Joanna Komorniczak}
\authorrunning{J. Komorniczak}
\institute{\textit{Department of Systems and Computer Networks},\\Wrocław University of Science and Technology}

\maketitle              

\begin{abstract}

Data streams are nowadays among the most frequently analyzed data structures, with the concept drift posing a major challenge encountered by processing systems. Despite the proposition of numerous solutions to counteract the accuracy degeneration due to concept drift, the scientific community has not yet established a unified framework for evaluating the concept drift detection task. Existing research often relies on classification quality metrics, but these can be affected by multiple factors and may not reliably reflect drift detection quality. In this work, we present an in-depth overview of the relationship between metrics for quantifying drift detection quality and classification performance in synthetic nonstationary data streams. The proposed research studies eight drift detection quality metrics in relation to the classifier's performance across seven synthetic data stream generation tools, additionally considering drift dynamics as a factor. The studies aim to identify the most informative set of drift detection quality metrics and provide a deep understanding of the method's evaluation.


\end{abstract}

\section{Introduction}
Most of the solutions employed in real-world systems rely on real-time inference, continuous system monitoring, and incremental model training. Combined with the velocity and volume of the processed data, this makes the data stream a~structure that is characteristic of modern machine learning systems~\cite{paramesha2024big}.

Data stream processing poses meaningful challenges related to the requirement of efficiency and accuracy of employed algorithms in a setting where the incrementally arriving data instances are sampled from distributions that  change over time~\cite{domingos2000mining}. Those underlying distribution changes are known as \textit{concept drifts} and may cause a decrease in the recognition quality of machine learning models~\cite{lu2018learning}.

\begin{figure}[!htb]
    \centering
    \includegraphics[width=\linewidth]{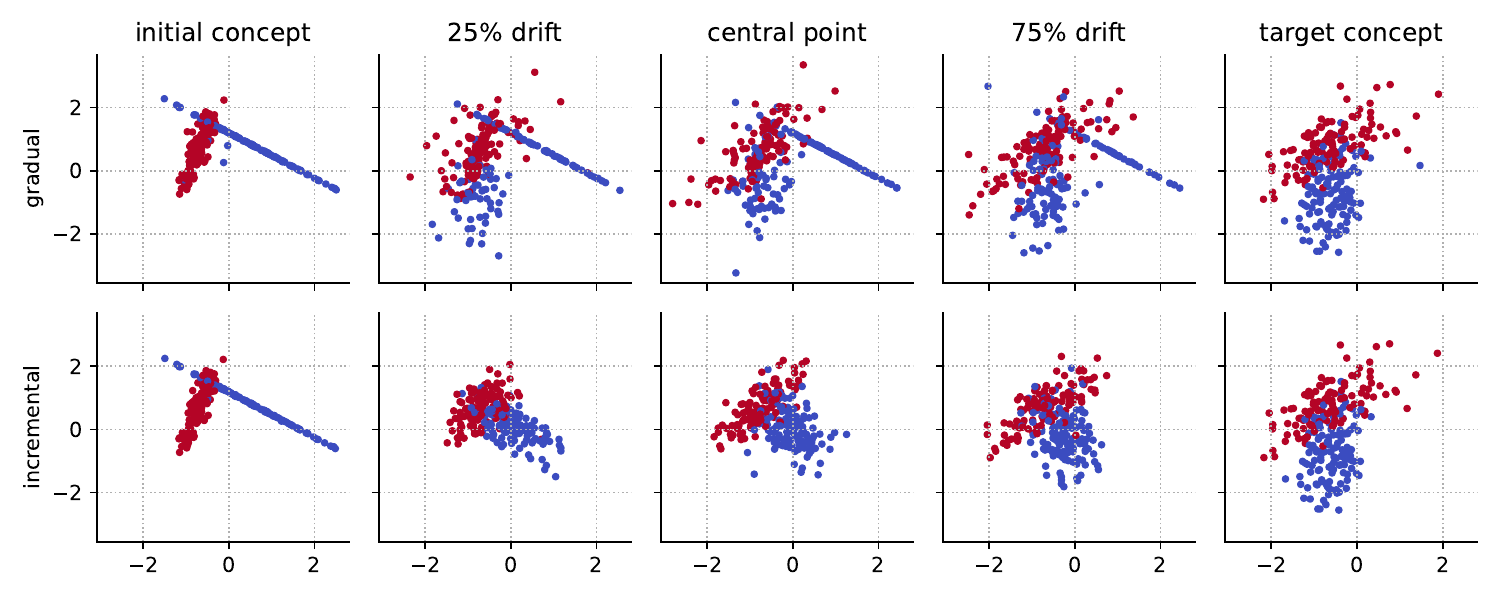}
    \caption{The example of gradual (top row) and incremental (bottom row) concept drift in a synthetic two-dimensional data stream}
    \label{fig:gradual-incremetnal}
\end{figure}

The concept drifts that impact recognition quality -- and therefore change the decision boundary of a classification problem -- are described as \textit{real} drifts. In the case of a \textit{virtual} drift, the data distribution changes, but it is not yet related to the loss of model performance~\cite{agrahari2022concept}. In terms of drift dynamics, concept drifts can be categorized as \textit{sudden}, \textit{gradual}, or \textit{incremental}. The sudden changes are often a result of an abrupt event that affected the system, such as sensor failure or replacement of its components~\cite{hinder2024one}. Gradual and incremental changes are more common in environments where change is observed over a longer time period, as is typical in social media and e-commerce, where trends change, and new solutions emerge~\cite{kolajo2019big}. In gradual drift, the data batch is a mixture of the old and a new concept, whereas in incremental change, the instances are drawn from a temporary distribution that is a superposition of the old and an incoming concept~\cite{suarez2023survey}. The examples of gradual and incremental drifts are presented in Figure~\ref{fig:gradual-incremetnal}.

To counteract the negative consequences of \textit{real} concept drift, machine learning solutions employ mechanisms of continuous or active adaptation~\cite{lin2025hybrid}. Such hybrid models often rely on concept drift detection mechanisms, formulated as a~separate task, where methods aim to mark the moments of concept changes~\cite{agrahari2022concept}.

Reliable evaluation of all machine learning solutions is of critical importance. The static-data-based evaluation protocol for recognition methods, such as classifiers and regressors, was adopted for data streams, resulting in the frequent use of prequential, test-then-train, or distributed cross-validation~\cite{bifet2015efficient} protocols in experiments. Meanwhile, the evaluation of the concept drift detection task is more complex -- since the detection timestamps need to be compared with the \textit{concept drift ground truth}~\cite{bifet2017classifier}, which is unavailable in real-world data. Often, researchers use classification quality as a \textit{proxy} for concept drift detection quality. Some research has shown that frequent and redundant detections may improve classification quality~\cite{liu2022concept}. However, other sources show that too many detections followed by classifier rebuilds may cause the model to forget past, yet important knowledge, resulting in a decrease in accuracy, even if the concept drift is correctly recognized~\cite{aguiar2024comprehensive}. These contrastive observations show that evaluating concept drift detection should not rely on classification quality but use dedicated metrics that evaluate the detection and drift timestamps. The necessity of concept drift ground truth makes the task rely almost entirely on synthetic data streams \cite{bifet2017classifier}.

\paragraph{Contribution}

The presented work aims to provide a response to the following \textit{research questions}:
\begin{itemize}
    \item \textbf{RQ1:} What is the relationship between various drift detection quality measures, and what is the correlation between classification accuracy and drift detection quality expressed by those dedicated metrics?
    \item \textbf{RQ2:} Which quality metrics are the most informative?
    \item \textbf{RQ3:} Does the selection of the synthetic data stream generation mechanism impact the quality assessment results?
    \item \textbf{RQ4:} Does the concept drift dynamics impact the quality assessment results?
\end{itemize}

To answer those questions, we design, conduct, and analyze four experiments on synthetic data streams. The main objective of this research is to study the experimental environment in which many classification and drift detection algorithms are evaluated. Therefore, the provided experiments do not compare existing drift detection solutions but rely on an abstract drift detector that randomly signals concept drifts with a specified detection probability. The evaluation uses a diverse set of synthetic data streams with various drift frequencies (Experiments 1 and 2), seven different generation mechanisms (Experiment 3), as well as three different drift dynamic types (Experiment 4).

\section{Related works}


Real concept drift negatively impacts the classifier's performance. Recognition methods aiming to minimize such impact often rely on ensembles, which can be categorized into \textit{active} or \textit{passive} approaches~\cite{agrahari2022concept}. Active ensembles employ a concept drift detection mechanism, which, in \textit{explicit} drift detectors, is based on classification quality~\cite{gozuaccik2021concept}. The canonical drift detectors, such as DDM and ADWIN, signaled detection directly based on the classifier's error rate~\cite{gama2004learning,bifet2007learning}. This evolution of drift detection methods may have been a factor in the still frequently employed strategy of relying on recognition quality to compare various algorithms. 
Meanwhile, modern drift detection solutions have shifted from accuracy-based data monitoring to unsupervised or implicit drift detection~\cite{lukats2025benchmark}, which also considers the limited availability of labels for real-time method evaluation~\cite{gomes2022survey}. In this challenging setting, studies have adopted metrics designed for the unbiased evaluation of the drift detection task.

\subsection{Evaluation of drift detection quality}\label{dd-metrics}

Many works rely on the classification quality to evaluate the effectiveness of the drift detection task~\cite{lu2018learning,gozuaccik2021concept}. In such an evaluation framework, the classifier is rebuilt each time the detection method marks a concept change, as is typical for \textit{active} ensembles.

Such an evaluation approach, however, has been shown not to reliably reflect the detection quality. A work by Bifet~\cite{bifet2017classifier} showed that rebuilding a classifier every fixed number of samples leads to better classification performance than all drift detection methods evaluated in the study. Therefore, oversensitive methods can obtain higher metric scores, as confirmed by Liu et al.~\cite{liu2022concept}. In a data stream setting, where system efficiency is extremely important, unnecessary model rebuilding and label acquisition will increase the cost of resources.

A recommended approach for drift detection evaluation is the use of dedicated quality metrics, which require the availability of exact concept drift moments \cite{lukats2025benchmark} and, therefore the use of synthetic data streams~\cite{souza2020challenges}.

Existing sources recommend calculating the \textit{true detection rate} (true positives), \textit{false detection rate} (false positives), and \textit{missed detection rate} (false negatives)~\cite{lu2018learning}. False positives, or false alarms, are also known as type I errors, and false negatives are referred to as type II errors~\cite{hinder2024one}. Some works parameterize those metrics with an acceptable delay~\cite{pesaranghader2016fast}, or a range of timestamps in which a drift is treated as a true positive~\cite{aguiar2024comprehensive}. Alternatively, in addition to a binary decision on whether a drift was correctly recognized, some authors quantify the time it took for the method to detect a drift~\cite{liu2022concept}. According to the timestamps of drifts and detections, five measures can be calculated~\cite{gustafsson2000adaptive,bifet2017classifier}:
\begin{itemize}
    \item Missed Detection Rate (MDR) -- quantifies the false negatives -- i.e., the drifts that were not recognized by the detector. The lower measure indicates better quality.
    \item Mean Time to Detection (MTD) -- a measure that quantifies the delay of drift detection. It considers only true positives to calculate the distance from drift to detection. Similar to MDR, the lower value indicates better quality.
    \item Mean Time between False Alarms (MTFA) -- considers false positives and measures the distance between them. At least two false alarms must occur in order to calculate MTFA. The higher measure indicates rare false alarms, therefore, a higher value is desired.
    \item Mean Time Ratio (MTR) is an aggregated measure that combines MDR, MTD, and MTFA. The goal is to maximize this measure.
    \item False Alarm Rate (FAR) -- calculated as the inverse of MTFA. Therefore, a good drift detector will obtain a low FAR.
\end{itemize}

When the method does not detect true drift, MDR and MTD will rise, indicating the worst detector operation. In contrast, when the drift detector is overreactive, MTFA and FAR measures will be impacted by a high number of false positives. One different measure that directly considers the number of detections compared to the number of drifts, and penalizes the detector for both too many and too few changes, is an \textit{R} measure, defined as part of \textit{drift detection error} measures~\cite{komorniczak2022statistical}:
\begin{itemize}
    \item The average distance of each detection to the nearest drift (D1).
    \item The average distance of each drift to the nearest detection (D2).
    \item The adjusted ratio of the number of drifts to the number of detections (R). This measure is scaled to obtain an optimal value of 0, the same as D1 and~D2.
\end{itemize}
Those parameter-less measures aim to describe the operation of a drift detector regardless of the drift dynamics. D1 considers both \textit{true positives} and \textit{false positives}, and will indicate a higher error when there is a significant number of detections that lie far from a concept drift. D2 seeks the nearest \textit{true positive} and quantifies the delay. The R measure focuses solely on the number of detections. This measure will indicate values above 1 if the number of detections is fewer than the number of actual drifts. 

The research community also proposed \textit{recovery measures}~\cite{shaker2015recovery} that rely on accuracy and are designed to quantify how well a recognition method handles nonstationary concepts. While the quality of drift detection can affect \textit{restoration time} and \textit{accuracy loss}, as with basic accuracy, these measures can be influenced by the data itself and may not reliably reflect the quality of drift detection.

\subsection{Synthetic generators}\label{synthetic-generators}

The drift detection quality evaluation requires the concept drift ground truth, which indicates the exact moments of concept changes. Therefore, the experiments on drift detection almost always rely on synthetic data streams, where those time instants are known and configurable.

The commonly used resources for data stream generation include the MOA, River, and scikit-multiflow \cite{montiel2018scikit} packages, as well as the generators that rely on Madelon datasets, available in the stream-learn library~\cite{ksieniewicz2022stream} -- where the data distribution characteristic and the informativeness of features can be precisely calibrated~\cite{guyon2004result}. Some generators, like OWDSG~\cite{komorniczak2025synthetic} and SynNC~\cite{masud2010classification}, can introduce novel classes in addition to concept drift. Each of those generators has its unique characteristics.

\begin{itemize}
\item \textit{SEA}~\cite{street2001streaming} -- synthesizes a binary classification problem described by three numerical features, only two of which are informative. The class is determined according to the sum of the informative attributes and a selected labeling function with a predetermined threshold. By default, the classes are imbalanced. The simulation of sudden concept drifts results from a random change of the labeling function.
\item \textit{Hyperplane}~\cite{hulten2001mining} -- binary classification problem described by 10 numerical features. The class of each instance is determined by the hyperplane equation with randomly generated weights. The classes are balanced.
\item \textit{RandomRBF}~\cite{montiel2018scikit} -- binary classification problem describing samples in 10 dimensions, where the points surround 50 centroids. The samples are drawn from a normal distribution with the expected value at the centroid. Clusters are assigned to classes randomly, and the problem is balanced.
\item \textit{Sine}~\cite{gama2004learning} -- binary classification data described by four attributes, two of which are informative. Features take values in the range from 0 to 1, and the class is determined based on the sine function. The problem is imbalanced.
\item \textit{AGRAWAL}~\cite{agrawal1993database} -- a generator describing a binary classification problem where a label determines the decision of loan approval. Samples are described by nine features, six of which are numerical and three are discrete. The problem is imbalanced. The generator implements 10 different functions that map features to synthetic labels.
\item \textit{StreamLearn}~\cite{ksieniewicz2022stream} -- a generator using the Madelon static data generator~\cite{guyon2004result}, in which class distributions are placed on the vertices of a multidimensional hypercube. The generated data presents a binary problem with two clusters per class, described by 20 numerical features, two of which are informative. By default, the classes are balanced. Concept drift involves the emergence of a new underlying distribution or a smooth transition between subsequent distributions in the case of gradual, incremental drift.
\item \textit{OWDSG}~\cite{komorniczak2025synthetic} -- similarly to the previous generator, data distributions are sampled using the Madelon procedure. The data describe a binary problem in 10 dimensions, where all features are informative. Each class is represented with points from a single cluster. Drift is simulated by changing the mask that defines the data cluster from which the class is sampled. The problems are balanced, and drifts are abrupt.
\end{itemize}

While those generators produce synthetic concepts that may not reliably represent real-world data, they are frequently employed in research due to the availability of configuring the moments of concept changes. Another benefit of using synthetic data streams is the ability to replicate the research using data generated with different random seeds.

\section{Experiment design}

The experiments were implemented in Python using the scikit-learn, stream-learn~\cite{ksieniewicz2022stream}, and scikit-multiflow~\cite{montiel2018scikit} libraries. The classification and regression algorithms, as well as the data stream generation methods, come from those libraries. All data streams were processed in equal-sized batches.

\subsection{Data streams}

The experiments used all seven generators described in Section~\ref{synthetic-generators}. In the Hyperplane, RandomRBF, Sine, and Agrawal streams, sudden concept drifts were simulated by reinitializing the generator with a new random seed at the moment of drift injection. SEA, StreamLearn, and OWDSG used their dedicated functions to inject a concept drift. 

All generated streams had 500 chunks, each containing 200 instances. In Experiment 2, only OWDSG was used; in Experiment 3, all generators were used with their default hyperparameters; and in Experiment 4, only StreamLearn was used to facilitate the comparison between various drift dynamics without changing the internal drift injection mechanism.

\subsection{Evaluation measures}

The experiments evaluated drift detection quality measures, classification accuracy, and their correlation. For drift detection evaluation, all eight metrics described in Section~\ref{dd-metrics} were used. To measure classification performance, accuracy (ACC) was calculated in cases of balanced data streams, and balanced accuracy (BAC) in cases where there was a possibility of a class imbalance.

To measure the correlation and associations between quality metrics, the research relied on the following measures:
\begin{itemize}
    \item Pearson Correlation Coefficient (PCC) -- measures the linear relationship between two sets of observations. It produces values from -1 to 1, with 0 indicating no correlation, values above 0 indicating a positive correlation, and values below 0 indicating a negative correlation.
    \item Spearman Correlation Coefficient (SCC) -- measures the monotonic relationship between sets of observations. Similar to the previous measure, it ranges from -1 to 1. In contrast to the PCC, it uses rank-order, therefore, it can capture non-linear but monotonic associations. 
    \item Mutual information (MI)~\cite{kraskov2004estimating} -- measures the linear and non-linear dependencies between sets of observations. It uses distances to k-nearest neighbors to estimate the value. It is equal to 0 when the observations are independent and takes larger values when the variables show relationship.  
\end{itemize}

\subsection{Goals of experiments}

\paragraph{Experiment 1} The first experiment focused on the evaluation of the relationship between drift detection error measures. The study used 8 measures, which were calculated between 1000 randomly drawn drift timestamps and detection timestamps. The probability of drift and detection occurrence was set at 20 different levels, ranging from 1\% to 15\% chance of an event. This resulted in 20,000 comparisons. Among replications, the probabilities for drift and detection occurrence were consistent. The number of data chunks in this synthetic setting was set to 1000. 

\paragraph{Experiment 2} The following experiment used synthetic data streams to evaluate classification performance and study the relationship between drift detection quality metrics and classification accuracy. Balanced binary classification data streams were generated using OWDSG. The streams described 500 chunks of size 200, where samples were described with 20 informative features. The streams in this experiment had 10 different drift numbers, ranging from 3 to 25, which corresponded to detection probabilities of 0.7\% to 5\%. All changes were sudden and uniformly distributed throughout the stream. The experiment was replicated 50 times for each of the 10 drift detection probabilities, resulting in the evaluation of 500 data streams. 

The experiment evaluated 5 classifiers: Gaussian Naive Bayes, Decision Tree, K-Nearest Neighbors, Support Vector Machine, and Multilayer Perceptron with their default configuration. Each model was fitted with the first data chunk and later used only to infer for the following batches. If the abstract detector signaled a concept change, the classifier was rebuilt and replaced the previous instance. Since the data streams were balanced, the experiment focused on the accuracy metric to evaluate the classification quality.

\paragraph{Experiment 2 Metaanalysis} The next experiment used the results of the previous one in Regression Metaanalysis. It aimed to examine the relationship between drift detection quality metrics and average classification accuracy. The experiment used a Multilayer Perceptron Regressor to determine which measures can be effectively estimated from the remaining ones, and reveal the most informative ones. The experiment employed 5-times repeated 2-fold cross-validation to examine the data from Experiment 2. All measures obtained in Experiment 2 were normalized. To evaluate the regression quality, the experiment used the Mean Squared Error (MSE) metric.

\paragraph{Experiment 3} The following experiment aimed to investigate the impact of the data stream generation mechanism on the relationship between drift detection quality and classification quality. Since some of the generators used in this research produce imbalanced data, the classification quality metrics were switched to balanced accuracy. As in Experiment 2, 500 data streams were generated (50 replications and 10 drift cardinalities and detection probabilities). This experiment evaluated all 7 data stream generation methods.

The previous experiment showed a consistent correlation pattern regardless of the classification algorithm, therefore, in this experiment, only an efficient GNB was used.

\paragraph{Experiment 4} The final experiment focused on evaluating the impact of drift dynamics on the correlation between detection quality and accuracy. The StreamLearn generator was used to produce data streams with sudden, gradual, and incremental drifts, along with 10 different drift numbers (from 3 to 10 drifts over 500 chunks). A concept sigmoid spacing parameter in the generator was set to 5 for gradual and incremental drifts. Stream generation for each configuration was replicated 50 times, resulting in the analysis of 1500 data streams. This experiment also used GNB classifier.

\section{Experiment results}

This section presents and analyzes the results of the performed experiments.

\subsection{Experiment 1 -- The correlation of drift detection quality metrics}

\begin{figure}[!htb]
    \centering
    \includegraphics[width=\linewidth]{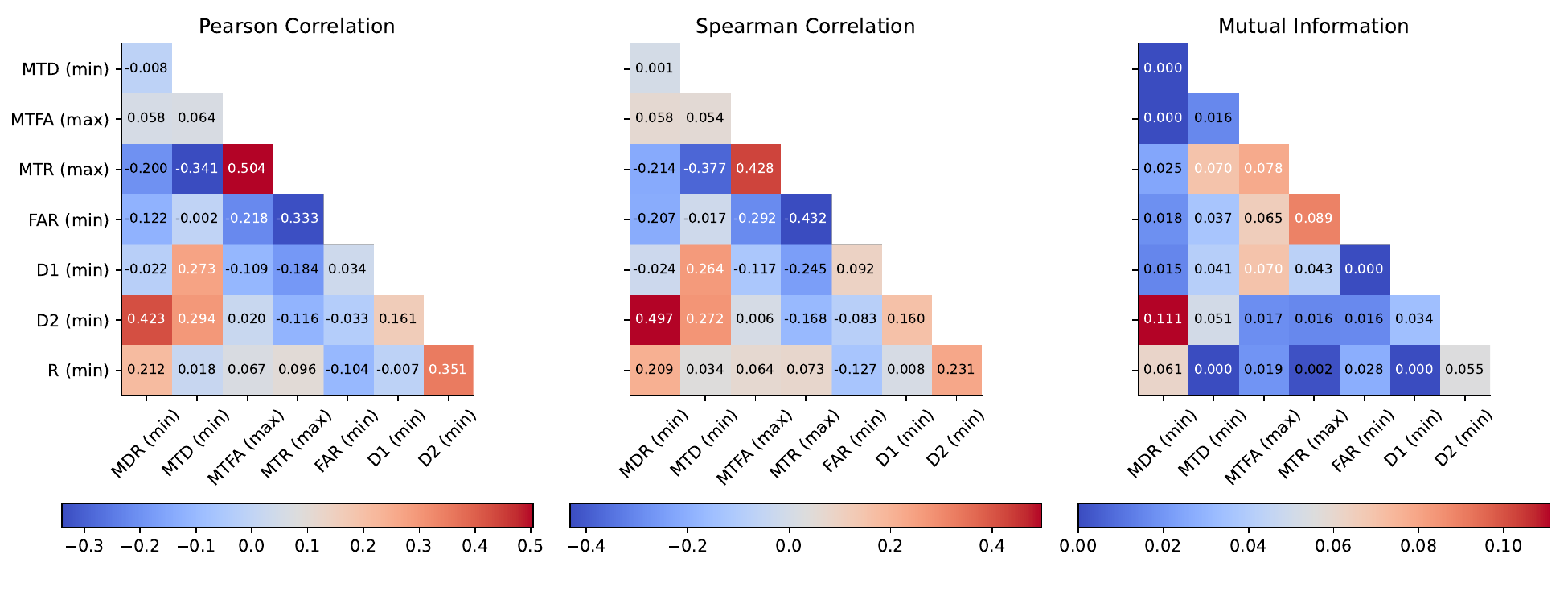}
    \caption{The measures studying the relationship between drift detection error measures: the PCC (left), SCC (center), and MI (right). The cells indicate the exact value of a relationship between specific metrics, shown in columns and in rows. The low correlation coefficient or mutual information is indicated with blue, and a high value with red.}
    \label{fig:e1}
\end{figure}

The first experiment studied the relationships between eight drift detection quality measures calculated based on timestamps of detections and a concept drift ground truth. The results are presented in Figure~\ref{fig:e1}.

The results show that the absolute values of the correlation coefficients are reaching up to approximately 0.5, indicating that the correlations exist but are not significant. MI indicates the strongest association between D2 and MDR, resulting from an increase in both of those metrics in the case of many redundant detections. The relation between these measures may be non-linear since the PCC for this metric pair is not the highest. MI also indicates a dependence between FAR and MTR. This results from both of these measures being dependent on MTFA. For this pair of measures, PCC and SCC indicate negative correlations. In contrast, correlation coefficients for MTR and MTFA are positive, resulting directly from the calculation of MTR based on MTFA. A measure that shows very little correlation (indicated by beige cells in the PCC and SCC heatmaps and blue in the MI heatmap) is the R measure. This results from the definition of this metric considering only the number of drifts and detections and not relying on the distances between the events.

\subsection{Experiment 2 -- The correlation of drift detection quality and classification accuracy}

The second experiment extended the previous evaluation with respect to classification accuracy. In the experiment, the classifier was rebuilt when the detection method marked a concept change. The research used 5 diverse classifiers, which achieved different classification qualities. Despite the variability in performance, the relationship between accuracy and the drift detection measures remained consistent, regardless of the base learner. For this reason, we present and analyze the results of the GNB classifier.

\begin{figure}[!t]
    \centering
    \includegraphics[width=\linewidth]{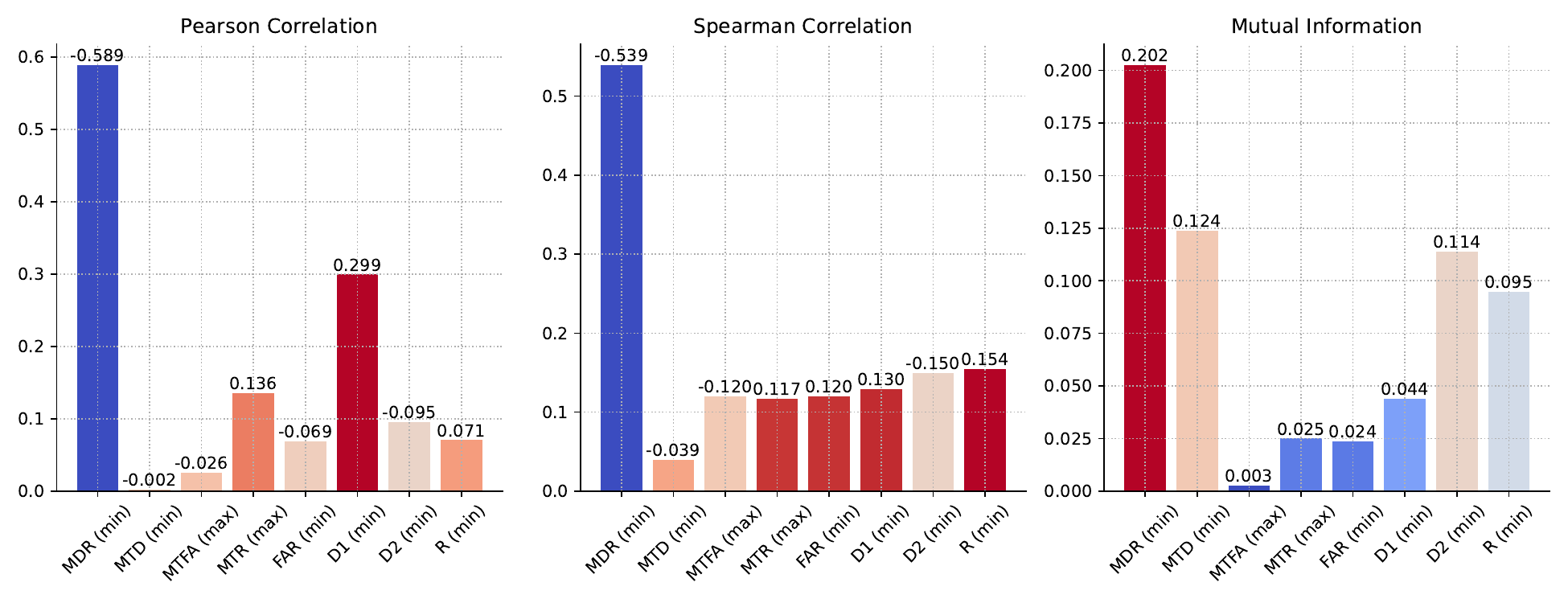}
    \caption{The relationship between specific drift detection measures (x-axis) and the classification accuracy, expressed in PCC (left), SCC (center), and MI (right). The height of a bar indicates the absolute value of a measure. Blue color indicates low value, and red color -- a high value.}
    \label{fig:e2}
\end{figure}

Figure~\ref{fig:e2} shows the results of this experiment. The height of a bar indicates the absolute measure value. To recognize the negative correlation in the case of PCC and SCC, the blue color was used. The closer the bar color is to red, the higher the correlation or the stronger the relationship (in the case of MI).

The strongest correlation is visible for MDR in terms of all three metrics. PCC and SCC show that the correlation is negative -- hence, higher accuracy results in lower MDR. This results from the impact of concept changes on the decision boundary, as failing to recognize the drift leads to a drop in accuracy. Since we aim for high accuracy and low MDR, these two optimization directions logically bond. Interestingly, all other drift detection metrics show a positive correlation, but out of the remaining measures, we only aim to maximize MTFA and MTR. This means that, even though the relation is not significant, when accuracy increases, FAR, D1, D2, and R also increase, indicating the worst result. MTD is an interesting measure since only MI reveals the dependence between its value and accuracy. We can therefore conclude that this relationship is non-linear.

The results of this experiment show that accuracy may not be a reliable assessment criterion since increasing accuracy may lead to a large number of false alarms (FAR) and a large distance from detections to the drift (D1).

\subsection{Metaanalysis of Experiment 2 -- Study of metric informativeness}

\begin{figure}[!htb]
    \centering
    \includegraphics[width=\linewidth]{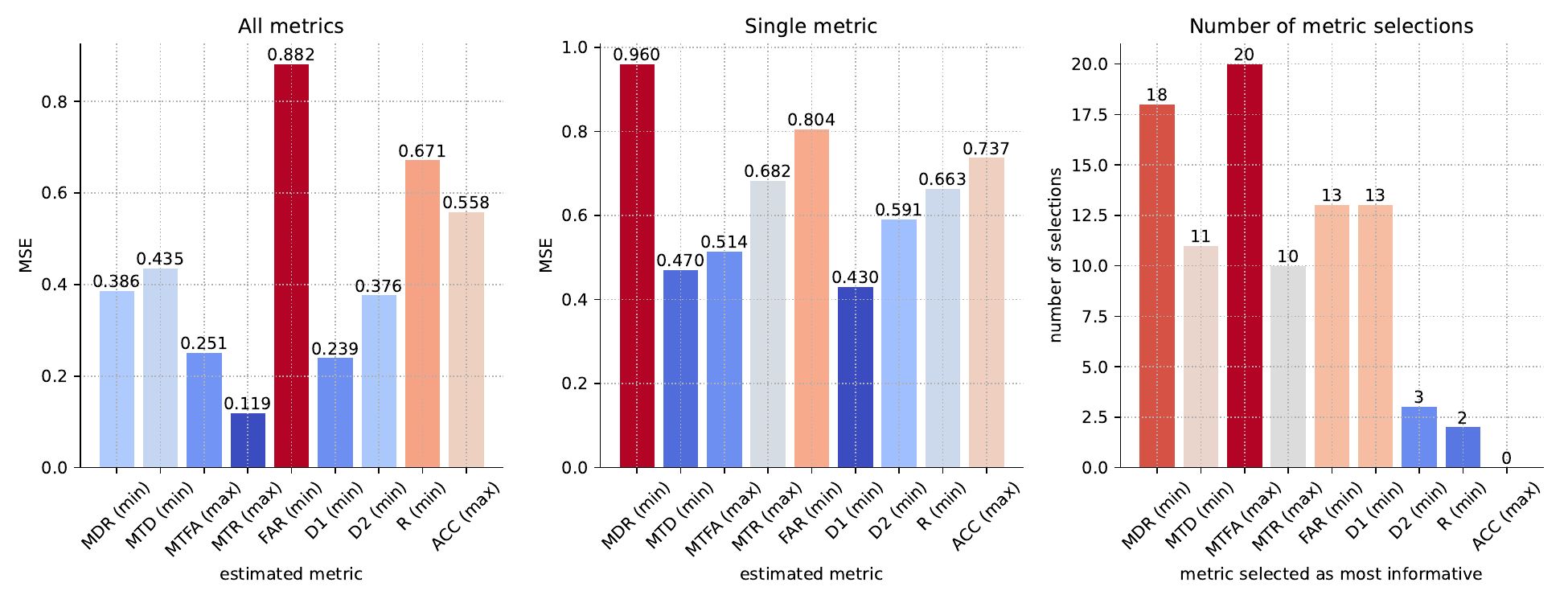}
    \caption{The MSE of the regression task, when aiming to estimate the specific measure using all remaining ones (left), using a single most informative metric (center), and the number of times the specific measure was selected as the most informative (right).}
    \label{fig:e2-meta}
\end{figure}

The following study is a metaanalysis of the results from Experiment 2, aiming to reveal the informativeness of each drift detection metric and indicate which may appear irrelevant. The results of this experiment are presented in Figure~\ref{fig:e2-meta}, indicating the MSE of the regression task when estimating a specific measure.

In the first part of the analysis, all remaining measures were used as features to estimate the target metric (indicated on the x-axis). The highest MSE is observed for FAR, R, and ACC. The number of false alarms is impacting both FAR and R, which was also observed in the first experiment. Since estimating their value is difficult, we can conclude that those measures add information in the evaluation of drift detection quality. A high error value for accuracy indicates that the exact estimation of its value is a challenge. Since the data streams were generated with various random seeds, some concepts may depict simpler distributions. Some of the injected concept drifts may also be virtual and may not cause a reduction in classification performance. The low MSE for MTR, D1, and MFTA indicates that those measures can be easily estimated using the remaining ones.

In the second part of the experiment, we selected the single most informative metric for each target. The selection was done using mutual information as a criterion~\cite{kraskov2004estimating}. As presented in the central part of Figure~\ref{fig:e2-meta}, the MDR exhibits the highest error, indicating that its value is difficult to estimate using a single other metric. MRT also obtained a higher error compared to the first part of the study, which may arise from the calculation of an MTR using three other measures. The right part of the figure shows how many times a specific measure was selected as the best descriptor. A high result for MTFA and MDR indicate their ability to approximate other measures. However, selecting a measure that has not been identified as a good \textit{proxy} may allow for the consideration of additional information to the experiments. The R measure, tied directly to the number of detections, was selected just 2 times as the estimation criterion. Meanwhile, we know that a low number of detections is valuable in terms of the costs of model rebuild and label acquisition. We also note that accuracy was not selected as an informative criterion, although its estimation error was not the highest. This may indicate that drift detection quality measures are more strongly bonded as a group than to accuracy.

\subsection{Experiment 3 -- The impact of generation mechanism}

\begin{figure}[!htb]
    \centering
    \includegraphics[width=\linewidth]{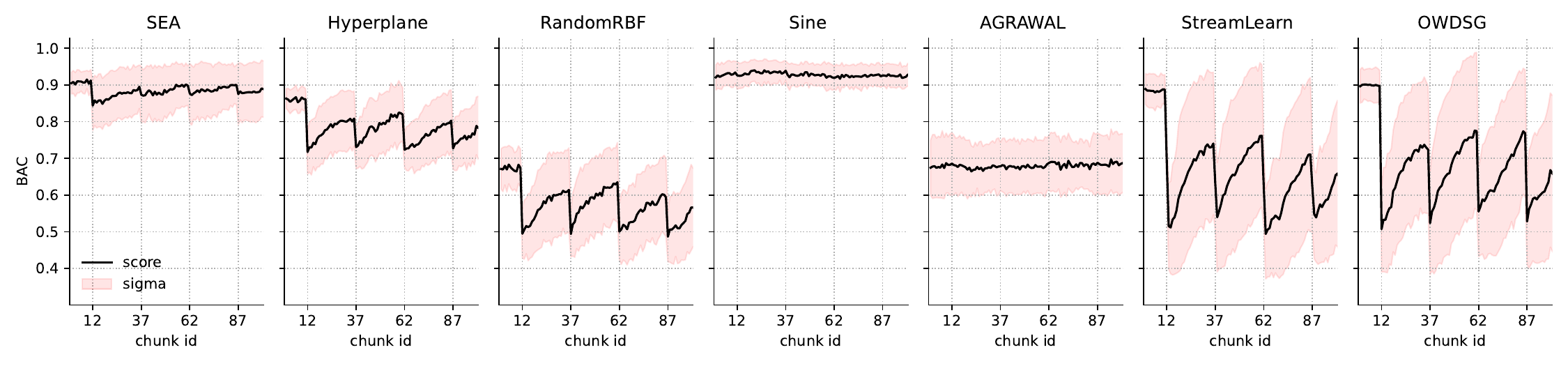}
    \caption{Averaged balanced accuracy of the classifier, when processing the initial 100 chunk of the data stream with four concept drifts visible. The black plot shows the average metric, and the red area indicates the standard deviation across replications.}
    \label{fig:e3_acc}
\end{figure}

The next experiment evaluated the data streams from seven generators and used balanced accuracy to measure the classification performance. First, we visualized how the metric changes due to concept drift. Figure~\ref{fig:e3_acc} shows the average impact in the first 100 data chunks of each stream.

Since the drift detection moments were random but related to the true number of drifts, the averaged results show a slow increase in classification performance after the initial drop in quality right after the drift. The results show that the generators produce data streams with various difficulty levels. Some streams (e.g. RandomRBF and AGRAWAL) pose a more significant challenge. Interestingly, the classification quality in the case of some generators, despite the concept drifts, remains relatively consistent. The Sine and AGRAWAL generators are examples of such. In SEA streams, the drifts impact the BAC score, but the changes are not as clearly visible as in Hyperplane, RandomRBF, StreamLearn, and OWDSG. The last two generators show the most significant drops in BAC, which is related to the possibility of describing diverse concepts in up to 20-dimensional feature space. Therefore, the concept drift will likely cause a shift in the decision boundary.

\begin{figure}[!htb]
    \centering
    \includegraphics[width=\linewidth]{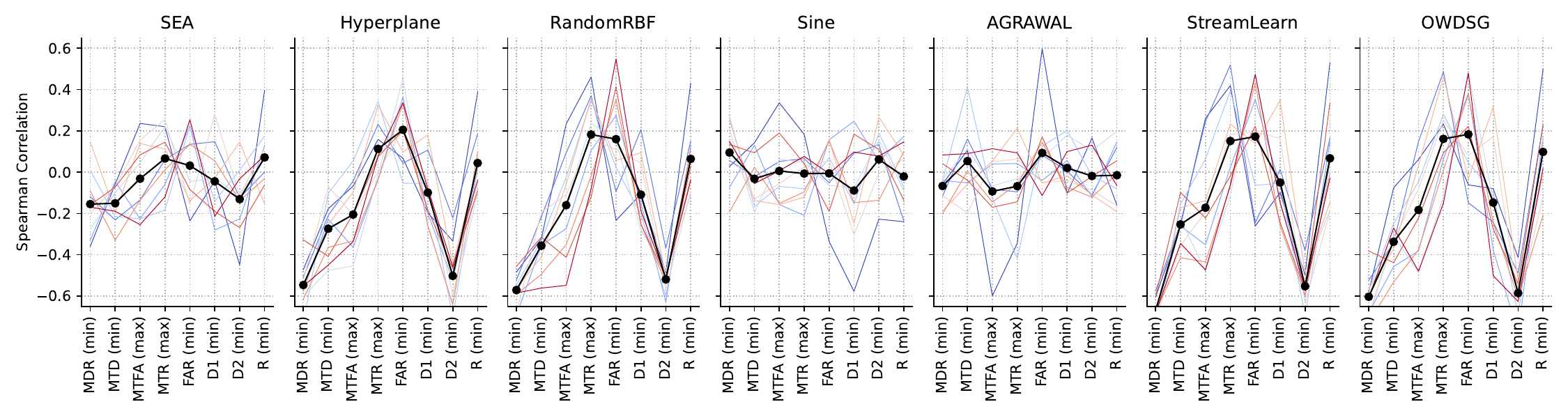}
    \caption{Data stream generator's fingerprints -- the SCC between BAC and drift detection quality metrics. The black line indicates the average regardless of the drift frequencies, and the colored lines indicate the correlations among a specific number of drifts.}
    \label{fig:e3_fingerprints}
\end{figure}

Figure~\ref{fig:e3_fingerprints} shows the \textit{fingerprints} of each generator, indicating the average SCC between BAC and drift detection quality metrics.

The figure shows a similar correlation pattern for Hyperplane, RandomRBF, StreamLearn, and OWDSG. The correlations in the case of those generators are stronger compared to SEA, Sine, and AGRAWAL. The significant correlation results from meaningful drops in BAC when the concept drift occurs. In the case of Sine and AGRAWAL, the concept drift caused minimal changes to BAC, therefore, the correlation is almost unnoticeable.

The lack of a correlation between BAC and drift detection quality may cause the evaluation of drift detection based on the classifier's performance to be even less related to the core of the task. Since it was previously shown that the frequent detections may result in a high classification performance~\cite{bifet2017classifier}, for some of the generated streams, this result may not provide much information about the drift detection quality.

\subsection{Experiment 4 -- How various drift dynamics change the correlation pattern}

\begin{figure}[!b]
    \centering
    \includegraphics[width=\linewidth]{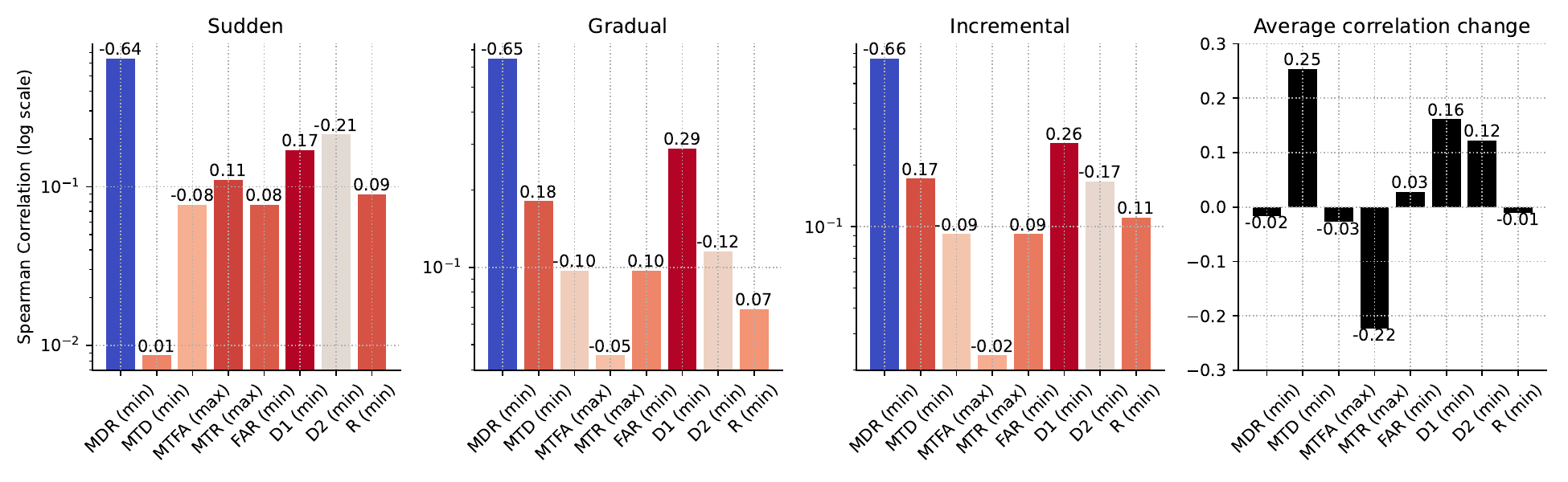}
    \caption{The SCC between specific drift detection metrics and accuracy in case of sudden, gradual, and incremental concept drift (first 3 subplots) and the difference in correlation (last subplot). The SCC plots use the logarithmic scale.}
    \label{fig:e4}
\end{figure}

The last experiment aimed to examine how various concept drift dynamics change the correlation patterns across measures. The results presented in Figure~\ref{fig:e4} show the SCC of drift detection metrics with accuracy for sudden, gradual, and incremental drifts. The last subplot shows the average change of correlations in the gradual and incremental concept drifts compared to sudden changes.

The results for gradual and incremental concept drifts show different patterns compared to sudden drifts. SCC in both cases is similar, since both are spread over time. If the detection is signaled during the transition period between concepts and the classifier is rebuilt, we would expect a rise in accuracy in both cases. Signaling concept drift multiple times over a single change could also positively impact accuracy, but it will negatively impact drift detection quality metrics.

The results show that the correlation for MDT increased, meaning that the worst result in this metric indicates better accuracy. This comes from the fact that, in non-sudden changes, the delayed detection is less impactful on classifier performance. The correlation with D1 and D2 also slightly increased. Marking many detections during the temporary concept will increase those error metrics, but also increase the classification accuracy. D1 and D2 were designed for the evaluation of all types of drifts since they quantify the distances to the central point of the drift. Therefore, they were expected to show more robustness to the impact of drift type compared to MTD. The results also show that the correlation with MTFA decreased to a value close to zero. In non-sudden changes, marking the single drift multiple times will improve classification quality but negatively impact this drift detection metric. The lack of correlation change in the case of MDR, FAR, and R comes from the fact that those metrics do not consider the delays or distances of detections to drifts, but only quantify how many changes have been recognized.

\section{Conclusions and Limitations}

The presented research aimed to provide insight into a reliable evaluation of the drift detection task in nonstationary data streams. The designed experiments studied the relationship between drift detection evaluation measures and their correlations with classification quality, which is frequently used as an evaluation proxy when the drift ground truth remains unknown. We summarize key findings from the performed experiments:
\begin{itemize}
    \item The experiments confirmed that accuracy is not a reliable proxy for drift detection quality, as high accuracy mainly indicates a low MDR.
    \item Some drift detection quality metrics (FAR, D1, D2, and R) indicate worse quality when the classification accuracy is high. Therefore, maximizing accuracy can lead to poor drift detection performance related to those metrics.
    \item Metaanalysis of results showed that MDR and MTFA provide the highest ability to approximate other metrics. The same experiment showed that the D2 and R metrics provide additional information about the detection quality.
    \item Different synthetic data stream generators exhibit different correlation \textit{fingerprints} between accuracy and drift detection quality metrics. Such correlation is consistently low in SEA, Sine, and AGRAWAL generators.
    \item The dynamics of concept drift impacts the results, increasing the correlation between MTD, D1, D2 and accuracy, and lowering the correlation between MTR and accuracy.
\end{itemize}

The work provided more insight into the drift detection evaluation, however, the study had certain limitations. The primary restriction was the inability to study metric relationships in real-world data, since the drift detection ground truth is required. Moreover, the experiments used the abstract drift detector, where the detection probability was tied to the number of drifts present in the stream. Meanwhile, the drift detection quality metrics are undefined if the detector fails to mark any change points or, as in the case of MTFA and MTR, when there are fewer than 2 false alarms. These limitations make evaluating the impact of drift detection in extreme cases impossible.

\begin{credits}
\subsubsection{\ackname} This work was supported by the statutory funds of the Department of Systems and Computer Networks, Faculty of Information and Communication Technology, Wrocław University of Science and Technology.

\subsubsection{\discintname}
The authors have no competing interests to declare that are relevant to the content of this article.
\end{credits}

\bibliographystyle{splncs04}
\bibliography{bib}

@inproceedings{bifet2015efficient,
  title={Efficient online evaluation of big data stream classifiers},
  author={Bifet, Albert and de Francisci Morales, Gianmarco and Read, Jesse and Holmes, Geoff and Pfahringer, Bernhard},
  booktitle={Proceedings of the 21th ACM SIGKDD international conference on knowledge discovery and data mining},
  pages={59--68},
  year={2015}
}

@article{agrahari2022concept,
  title={Concept drift detection in data stream mining: A literature review},
  author={Agrahari, Supriya and Singh, Anil Kumar},
  journal={Journal of King Saud University-Computer and Information Sciences},
  volume={34},
  number={10},
  pages={9523--9540},
  year={2022},
  publisher={Elsevier}
}

@article{lin2025hybrid,
  title={Hybrid ensemble framework for imbalanced data streams with concept drift},
  author={Lin, Mianfen and Yu, Zhiwen and Yang, Kaixiang and Chen, CL Philip},
  journal={IEEE Transactions on Big Data},
  year={2025},
  publisher={IEEE}
}

@article{suarez2023survey,
  title={A survey on machine learning for recurring concept drifting data streams},
  author={Su{\'a}rez-Cetrulo, Andr{\'e}s L and Quintana, David and Cervantes, Alejandro},
  journal={Expert Systems with Applications},
  volume={213},
  pages={118934},
  year={2023},
  publisher={Elsevier}
}

@article{hinder2024one,
  title={One or two things we know about concept drift—a survey on monitoring in evolving environments. Part A: detecting concept drift},
  author={Hinder, Fabian and Vaquet, Valerie and Hammer, Barbara},
  journal={Frontiers in Artificial Intelligence},
  volume={7},
  pages={1330257},
  year={2024},
  publisher={Frontiers Media SA}
}

@inproceedings{domingos2000mining,
  title={Mining high-speed data streams},
  author={Domingos, Pedro and Hulten, Geoff},
  booktitle={Proceedings of the sixth ACM SIGKDD international conference on Knowledge discovery and data mining},
  pages={71--80},
  year={2000}
}

@article{kraskov2004estimating,
  title={Estimating mutual information},
  author={Kraskov, Alexander and St{\"o}gbauer, Harald and Grassberger, Peter},
  journal={Physical Review E—Statistical, Nonlinear, and Soft Matter Physics},
  volume={69},
  number={6},
  pages={066138},
  year={2004},
  publisher={APS}
}

@article{masud2010classification,
  title={Classification and novel class detection in concept-drifting data streams under time constraints},
  author={Masud, Mohammad and Gao, Jing and Khan, Latifur and Han, Jiawei and Thuraisingham, Bhavani M},
  journal={IEEE Transactions on knowledge and data engineering},
  volume={23},
  number={6},
  pages={859--874},
  year={2010},
  publisher={IEEE}
}

@article{guyon2004result,
  title={Result analysis of the nips 2003 feature selection challenge},
  author={Guyon, Isabelle and Gunn, Steve and Ben-Hur, Asa and Dror, Gideon},
  journal={Advances in neural information processing systems},
  volume={17},
  year={2004}
}

@article{shaker2015recovery,
  title={Recovery analysis for adaptive learning from non-stationary data streams: Experimental design and case study},
  author={Shaker, Ammar and H{\"u}llermeier, Eyke},
  journal={Neurocomputing},
  volume={150},
  pages={250--264},
  year={2015},
  publisher={Elsevier}
}

@article{komorniczak2022statistical,
  title={Statistical drift detection ensemble for batch processing of data streams},
  author={Komorniczak, Joanna and Zyblewski, Pawe{\l} and Ksieniewicz, Pawe{\l}},
  journal={Knowledge-Based Systems},
  volume={252},
  pages={109380},
  year={2022},
  publisher={Elsevier}
}

@article{lukats2025benchmark,
  title={A benchmark and survey of fully unsupervised concept drift detectors on real-world data streams},
  author={Lukats, Daniel and Zielinski, Oliver and Hahn, Axel and Stahl, Frederic},
  journal={International Journal of Data Science and Analytics},
  volume={19},
  number={1},
  pages={1--31},
  year={2025},
  publisher={Springer}
}

@article{paramesha2024big,
  title={Big data analytics, artificial intelligence, machine learning, internet of things, and blockchain for enhanced business intelligence},
  author={Paramesha, Mallikarjuna and Rane, Nitin and Rane, Jayesh},
  journal={Artificial Intelligence, Machine Learning, Internet of Things, and Blockchain for Enhanced Business Intelligence (June 6, 2024)},
  year={2024}
}

@article{liu2022concept,
  title={Concept drift detection delay index},
  author={Liu, Anjin and Lu, Jie and Song, Yiliao and Xuan, Junyu and Zhang, Guangquan},
  journal={IEEE Transactions on Knowledge and Data Engineering},
  volume={35},
  number={5},
  pages={4585--4597},
  year={2022},
  publisher={IEEE}
}

@article{souza2020challenges,
  title={Challenges in benchmarking stream learning algorithms with real-world data},
  author={Souza, Vinicius MA and dos Reis, Denis M and Maletzke, Andre G and Batista, Gustavo EAPA},
  journal={Data Mining and Knowledge Discovery},
  volume={34},
  number={6},
  pages={1805--1858},
  year={2020},
  publisher={Springer}
}

@article{gomes2022survey,
  title={A survey on semi-supervised learning for delayed partially labelled data streams},
  author={Gomes, Heitor Murilo and Grzenda, Maciej and Mello, Rodrigo and Read, Jesse and Le Nguyen, Minh Huong and Bifet, Albert},
  journal={ACM Computing Surveys},
  volume={55},
  number={4},
  pages={1--42},
  year={2022},
  publisher={ACM New York, NY}
}

@inproceedings{bifet2017classifier,
  title={Classifier concept drift detection and the illusion of progress},
  author={Bifet, Albert},
  booktitle={International conference on artificial intelligence and soft computing},
  pages={715--725},
  year={2017},
  organization={Springer}
}

@article{aguiar2024comprehensive,
  title={A comprehensive analysis of concept drift locality in data streams},
  author={Aguiar, Gabriel J and Cano, Alberto},
  journal={Knowledge-Based Systems},
  volume={289},
  pages={111535},
  year={2024},
  publisher={Elsevier}
}

@inproceedings{pesaranghader2016fast,
  title={Fast hoeffding drift detection method for evolving data streams},
  author={Pesaranghader, Ali and Viktor, Herna L},
  booktitle={Joint European conference on machine learning and knowledge discovery in databases},
  pages={96--111},
  year={2016},
  organization={Springer}
}

@article{lu2018learning,
  title={Learning under concept drift: A review},
  author={Lu, Jie and Liu, Anjin and Dong, Fan and Gu, Feng and Gama, Joao and Zhang, Guangquan},
  journal={IEEE transactions on knowledge and data engineering},
  volume={31},
  number={12},
  pages={2346--2363},
  year={2018},
  publisher={IEEE}
}

@book{gustafsson2000adaptive,
  title={Adaptive filtering and change detection},
  author={Gustafsson, Fredrik and Gustafsson, Fredrik},
  volume={1},
  year={2000},
  publisher={Wiley New York}
}

@inproceedings{bifet2007learning,
  title={Learning from time-changing data with adaptive windowing},
  author={Bifet, Albert and Gavalda, Ricard},
  booktitle={Proceedings of the 2007 SIAM international conference on data mining},
  pages={443--448},
  year={2007},
  organization={SIAM}
}

@article{montiel2018scikit,
  title={Scikit-multiflow: A multi-output streaming framework},
  author={Montiel, Jacob and Read, Jesse and Bifet, Albert and Abdessalem, Talel},
  journal={Journal of Machine Learning Research},
  volume={19},
  number={72},
  pages={1--5},
  year={2018}
}

@article{gozuaccik2021concept,
  title={Concept learning using one-class classifiers for implicit drift detection in evolving data streams},
  author={G{\"o}z{\"u}a{\c{c}}{\i}k, {\"O}mer and Can, Fazli},
  journal={Artificial Intelligence Review},
  volume={54},
  number={5},
  pages={3725--3747},
  year={2021},
  publisher={Springer}
}

@article{kolajo2019big,
  title={Big data stream analysis: a systematic literature review},
  author={Kolajo, Taiwo and Daramola, Olawande and Adebiyi, Ayodele},
  journal={Journal of Big Data},
  volume={6},
  number={1},
  pages={1--30},
  year={2019},
  publisher={Springer}
}

@inproceedings{gama2004learning,
  title={Learning with drift detection},
  author={Gama, Joao and Medas, Pedro and Castillo, Gladys and Rodrigues, Pedro},
  booktitle={Brazilian symposium on artificial intelligence},
  pages={286--295},
  year={2004},
  organization={Springer}
}

@article{ksieniewicz2022stream,
  title={Stream-learn—open-source python library for difficult data stream batch analysis},
  author={Ksieniewicz, Pawel and Zyblewski, Pawel},
  journal={Neurocomputing},
  volume={478},
  pages={11--21},
  year={2022},
  publisher={Elsevier}
}

@inproceedings{komorniczak2025synthetic,
  title={Synthetic Non-stationary Data Streams for Recognition of the Unknown},
  author={Komorniczak, Joanna},
  booktitle={Joint European Conference on Machine Learning and Knowledge Discovery in Databases},
  pages={143--159},
  year={2025},
  organization={Springer}
}

@inproceedings{street2001streaming,
  title={A streaming ensemble algorithm (SEA) for large-scale classification},
  author={Street, W Nick and Kim, YongSeog},
  booktitle={Proceedings of the seventh ACM SIGKDD international conference on Knowledge discovery and data mining},
  pages={377--382},
  year={2001}
}

@inproceedings{hulten2001mining,
  title={Mining time-changing data streams},
  author={Hulten, Geoff and Spencer, Laurie and Domingos, Pedro},
  booktitle={Proceedings of the seventh ACM SIGKDD international conference on Knowledge discovery and data mining},
  pages={97--106},
  year={2001}
}

@article{agrawal1993database,
  title={Database mining: A performance perspective},
  author={Agrawal, Rakesh and Imielinski, Tomasz and Swami, Arun},
  journal={IEEE transactions on knowledge and data engineering},
  volume={5},
  number={6},
  pages={914--925},
  year={1993},
  publisher={IEEE}
}

\end{document}